\documentclass[11pt]{article}

\usepackage{acl}
\usepackage{times}
\usepackage{latexsym}
\usepackage[T1]{fontenc}
\usepackage[utf8]{inputenc}
\usepackage{microtype}
\usepackage{inconsolata}
\usepackage{graphicx}
\usepackage{booktabs}
\usepackage{multirow}
\usepackage{amsmath}
\usepackage{amssymb}
\usepackage{enumitem}
\usepackage{adjustbox}
\usepackage{float}
\usepackage{placeins}
\usepackage{tikz}
\usepackage{pgfplots}
\pgfplotsset{compat=1.18}
\usetikzlibrary{positioning}
\setlist[itemize]{topsep=3pt,itemsep=2pt,parsep=0pt,partopsep=0pt}

\title{Retention Consequence in Lifecycle Memory Control}

\author{Jiarui Han\\j22han@uwaterloo.ca}

\begin{document}
\maketitle

\begin{abstract}
Persistent memory can fail after successful admission: a premise is written, then becomes a silent assumption, and later maintenance treats it as ordinary residue to be compressed, demoted, or evicted. We study this post-admission failure as a lifecycle-control problem. Existing memory systems already perform admission, update, compression, retrieval, and eviction. Our claim is not that such systems lack maintenance, but that retention consequence is often operationalized only indirectly through validity, similarity, recency, frequency, importance, or summarization signals rather than exposed as a separate lifecycle state. We therefore treat \texttt{confidence} as carried-forward validity/support evidence, and introduce \texttt{strength} as an explicit lifecycle state for retention consequence. We operationalize this distinction in StageMem, a small staged controller whose transient, working, and durable stores expose promotion, compression, and eviction pressure points. Across controlled premise-realization, compression, pressure, and implicit-heuristic diagnostics, the experiments separate writing too little, retaining the wrong high-cue content, forgetting costly premises, and preserving everything by saturation. Explicit retention consequence, used through lifecycle settlement, provides a control surface between omission and hoarding. For the targeted post-admission failure mode, the results support a lifecycle view of persistent memory: reliability depends not only on what enters memory, but on whether admission validity and retention consequence remain available during maintenance.
\end{abstract}

\section{Introduction}

Persistent memory can fail after it succeeds. A project constraint, user preference, role constraint, or task assumption may be written early and then become a silent premise: the user stops repeating it precisely because it is expected to remain in force. Later maintenance may compress, demote, or evict it together with ordinary local residue. The failure is easy to miss: the model may still produce fluent, relevant-looking answers while its behavior is no longer governed by the assumption the user expected it to remember.

The issue is not only what enters memory, but what remains protected when memory becomes pressured. Existing memory systems already update, consolidate, retrieve, summarize, compress, and evict stored content, often using signals such as validity, similarity, recency, repetition, relevance, conflict status, or generic importance. These signals can rank content, but they tend to privilege what is currently supported, repeated, query-near, or locally salient. An early governing premise can therefore become weaker in the maintenance signal even as its behavioral consequence remains high. A repeated local detail, a query-near support fact, and a rare governing premise may all look useful under some heuristic, yet they should not occupy the same lifecycle status.

This is the gap we target. Retention consequence may correlate with familiar signals, but correlation is not control. When forgetting cost is only a side effect of validity, relevance, frequency, or summarization heuristics, costly premises are protected only when those heuristics happen to agree. We therefore treat persistent memory as a lifecycle-control problem: after admission, the system must still decide which items remain tentative, which become durable, which can be compressed, and which can be forgotten.

We operationalize this view with two carried state variables. \texttt{Confidence} carries forward admission-time evidence that an item is valid, supported, or still usable; \texttt{strength} records the retention consequence of losing it under later maintenance. The pairing matters because neither signal can do the other's job: without confidence, strength can protect false importance cues; without strength, confidence can preserve credible residue while losing the premise that mattered. Strength is therefore not a generic importance label, but a state used to decide how admitted content should survive later pressure.

We make this decomposition testable in StageMem, a compact transient/working/durable controller. The stages are not the contribution by themselves; they provide pressure points where the two signals can act. Newly admitted items begin as tentative memories, compete locally, and can become durable only when confidence and retention consequence justify deeper commitment. This gives forgetting cost process semantics rather than treating it as another flat score.

The experiments test this control claim rather than rank full memory products. Because existing long-memory benchmarks mainly test retrieval, relation traversal, coverage, or end-to-end dialogue behavior, we use controlled diagnostics as the evidential core: extraction, final readout, and metric definitions are held fixed while the post-admission control policy varies. Premise-realization, compression, pressure, gate--strength, confidence$\times$strength, strength-source, and implicit-heuristic diagnostics test whether confidence, strength, lifecycle staging, or familiar heuristics can substitute for one another. External adaptations then check compatibility with retrieval- and relation-heavy settings.

Our contributions are:
\begin{itemize}[leftmargin=*]
\item We identify \emph{post-admission premise loss} as a distinct failure mode for persistent memory: a fact can be admitted correctly yet later stop governing behavior because maintenance treats it as ordinary residue.
\item We formulate lifecycle memory control around two carried signals. \texttt{Confidence} carries forward admission-time validity/support evidence, while \texttt{strength} makes retention consequence explicit rather than leaving forgetting cost implicit inside relevance, recency, frequency, or generic importance heuristics.
\item We instantiate this formulation in StageMem, a compact transient/working/durable controller that makes promotion, compression, and eviction decisions inspectable under bounded capacity.
\item We provide controlled diagnostics showing that explicit retention consequence changes the failure surface rather than merely increasing recall, separating omission, false-cue protection, costly-premise loss, and saturation.
\end{itemize}

\section{Related Work}

\paragraph{Persistent and evolving memory.}
A growing line of work equips language-model systems with persistent memory across sessions and tasks. MemoryBank, Mem0, A-MEM, All-Mem, AtomMem, and reflective memory-management approaches study how memories are extracted, updated, consolidated, retrieved, and revised over time \citep{memorybank2023,mem02025,amem2025,allmem2026,atommem2026,reflective2025}. These systems establish memory as an evolving substrate rather than a static cache. We focus on a lower-level control question inside that substrate: once content has entered memory, what state determines whether losing it would be consequential? We separate carried validity/support evidence from retention consequence, so that forgetting cost is not operationalized only indirectly through extraction confidence, salience, recency, or consolidation heuristics.

\paragraph{Agent memory, reflection, and priority signals.}
Agent-memory systems further show how memory signals shape future behavior. Generative Agents store observations in a memory stream, retrieve records using recency, importance, and relevance, and synthesize reflections for later planning and interaction \citep{generativeagents2023}. Reflexion stores verbal feedback and self-reflections in an episodic memory buffer to improve subsequent agent decisions \citep{reflexion2023}. These systems make memory actionable, but their priority signals mainly guide retrieval, reflection, or behavioral improvement. Our question is different: whether an admitted item carries an explicit state for the consequence of being forgotten during later maintenance.

\paragraph{Hierarchical and lifecycle memory.}
Hierarchical memory systems such as MemGPT, HiMem, and H-MEM-style approaches organize content into levels, buffers, or stores \citep{memgpt2023,himem2026,hmem2025}. This supports the intuition that memory should not be a single undifferentiated store. Our contribution is complementary: we ask what item-level state should drive movement through such a lifecycle. In our formulation, hierarchy is useful because it creates pressure points where confidence and retention consequence can be acted on; the paper does not depend on a particular number of layers being uniquely optimal.

\paragraph{Long context, compression, and memory benchmarks.}
Long-context models and retrieval systems do not always use relevant information robustly across long inputs \citep{longmem2023,lostmiddle2024}. Prompt and context compression work treats compression as a lossy budgeted process that must preserve task-relevant information \citep{llmlingua2023,longllmlingua2024,selectivecontext2023}, while dialogue-summary evaluations show that summaries can omit or distort facts \citep{diac2024}. Long-memory benchmarks such as LoCoMo, LongMemEval, MemBench, and LoCCo broaden evaluation beyond short-context retrieval to long-term QA, summarization, temporal reasoning, knowledge updates, relation traversal, and long-document coverage \citep{locomo2024,longmemeval2024,membench2025,locco2025}. These benchmarks are valuable, but they do not directly isolate the maintenance failure studied here: an admitted item faces later maintenance, and final behavior must still be governed by the costly-to-forget premise. We therefore use controlled diagnostics for mechanism separation and external adaptations as compatibility checks.

\section{Method: Lifecycle Memory Control}

\subsection{Post-admission premise loss}

We isolate the part of memory control that begins after a candidate item is available to the memory system. The hard case is not a one-time write decision, but later competition under capacity pressure. Long-horizon premises are often introduced early as global instructions, task assumptions, role constraints, or project goals; users then stop repeating them because they assume the premise remains in force. Such premises are vulnerable to maintenance policies that favor recent, repeated, query-near, or locally salient content.

We call the target failure \emph{post-admission premise loss}. A premise counts as lost when later maintenance prevents it from governing the final readout, even if it was admitted earlier or related facts remain available. The loss may be gradual: an explicit premise is compressed into a weaker summary, then into a generic task note, and finally discarded as ordinary residue. The system may still preserve nearby details while losing the premise that determines how those details should be used.

We factor the pipeline to make this failure identifiable. Candidate proposal, symbolic encoding, final readout, and metrics are shared; the varied component is the post-admission control policy. The goal is not to measure extraction or generation quality, but to ask a narrower causal question: under the same admitted stream and budget pressure, which controller keeps costly premises actionable instead of letting them collapse into ordinary residue?

\subsection{Validity evidence and retention consequence}

The target failure arises when a premise remains true enough to use, but no longer strong enough to survive maintenance. This requires separating two quantities that ordinary priority scores can blur. An admitted item may be well supported, recent, repeated, or query-relevant, yet cheap to forget. Another item may be old or sparse in the stream, yet costly to lose because future behavior depends on it. We therefore separate the evidence that an item is usable from the consequence of losing it.

We use \texttt{confidence} for the first role. It carries forward the kind of validity/support evidence many systems already use at the front door---extraction confidence, local support, relevance, conflict status, or write-worthiness---so that later maintenance can still reject invalid, revoked, or weakly supported content. We use \texttt{strength} for the second role: how costly it would be for an admitted item to lose operational force under later pressure. Such a premise should not have to survive by remaining recent or repeated; it should survive because forgetting it would change what the system should do.

The coupled design follows from the failure modes we target. Strength without a carried validity gate can protect false high-cue content. Confidence without strength can preserve credible local residue while losing a costly premise. A flat or binary priority can preserve marked content but tends toward saturation or brittle tie-breaking under bounded capacity. The resulting controller must therefore combine a carried validity gate, an explicit retention-consequence state, and a settlement process that decides when these signals affect promotion, compression, and eviction.

\subsection{StageMem: a reference controller}

StageMem makes the decomposition operational by giving admitted items different lifecycle statuses. It is a compact reference controller for testing how carried validity evidence and retention consequence affect maintenance decisions under bounded capacity. The architectural idea is that forgetting cost should affect movement through a lifecycle, not only a one-shot ranking score.

The controller organizes memory into three stages:
\begin{itemize}[leftmargin=*]
\item \textbf{Transient memory}: low-commitment storage for newly admitted items.
\item \textbf{Working memory}: intermediate storage for items that survive local competition.
\item \textbf{Durable memory}: deep storage for content with enough support or consequence to justify long-term retention.
\end{itemize}

Staging gives retention consequence a place to act. A flat store can assign a priority to every item, but it does not by itself distinguish a tentative write from an active working assumption or a durable premise. StageMem makes these commitment levels explicit: new content is tested in transient memory, working memory supports local competition and update, and durable memory protects premises whose loss would change later behavior. When a stage fills, settlement decides whether items are discarded, retained, compressed, or promoted.

Settlement uses confidence as a validity gate and strength as retention consequence. In the current implementation, an item is eligible for promotion only when its confidence exceeds a stage-specific validity threshold and its strength exceeds a stage-specific retention threshold:
\begin{equation}
\mathrm{promote}(i)=\mathbb{1}[c_i\ge\tau_c]\cdot\mathbb{1}[r_i\ge\tau_r].
\end{equation}
Items that fail settlement can remain shallow, be compressed, or be evicted depending on the stage and budget. The exact admission heuristic, update equations, and threshold settings are implementation choices reported in the appendix; the main experiments test the control property that validity evidence and retention consequence are carried forward and acted on during maintenance.

\subsection{Updates and extensibility}

Confidence and strength are carried states, not fixed labels. They can be updated by different evidence: corroboration or contradiction changes whether an item is trustworthy, while later use, dependency, user pinning, or task structure changes how costly it would be to forget. This distinction prevents the two states from collapsing back into one score. A repeated local detail may become familiar without becoming durable; a rarely repeated premise may remain high-strength because later behavior depends on it.

The current implementation uses simple auditable update rules, with full equations and salience initialization reported in the appendix. Learned salience, cue rules, user feedback, graph-derived dependencies, or domain policies can initialize or update strength, but they do not replace lifecycle control: retention consequence must still be carried forward and acted on during promotion, compression, and eviction. Graph retrieval is complementary: it improves traversal among stored items, while StageMem specifies how admitted items are retained, weakened, or forgotten under maintenance pressure.

\section{Experimental Setup}

\subsection{Shared symbolic harness}

The experiments use a shared symbolic harness to isolate post-admission memory control. Unless otherwise noted, all controllers receive the same candidate stream, capacity budget, symbolic encoder, query interface, and metric definitions. Candidate proposal and final readout are fixed, so differences reflect the maintenance policy: what is retained, promoted, compressed, or evicted after admission.

The setting is intentionally diagnostic. The required premise/support chain either remains available for final symbolic readout or it does not, so 0/1 outcomes are interpreted as failure indicators rather than estimates of average product performance. The goal is to identify which control assumption fails under pressure: omission, premise loss, support loss, false protection, or saturation.

\subsection{Compared controllers}

We compare StageMem with component ablations and control-style analogues. Component ablations include confidence-only settlement, strength-only lifecycle control, single-state memory, single-layer memory, and variants that remove or collapse retention consequence. Control-style analogues instantiate broad memory-management principles under the shared harness: conservative front-door writing, reinforced flat storage, aggressive tiering, and hybrid layering. They are not product reproductions; they place different control principles under the same maintenance contract.

We also include stronger diagnostic controls for narrower objections. Cue-aware flat scoring, binary flags, and front-door sweeps use obvious importance cues to test whether ``just mark it important'' is enough under budget. The implicit-heuristic diagnostic tests whether recency, frequency, query relevance, generic importance, or cue-aware flat salience can substitute for explicit retention consequence.

\subsection{Metrics and failure axes}

The main metrics are premise survival (\textbf{Prem}), support realization (\textbf{Supp}), final behavior realization (\textbf{Behav}), false-premise protection (\textbf{FalsePrem}), final normalized load, noncritical residue, and budget-hit rate. \textbf{Prem} tracks whether the governing premise remains actionable; \textbf{Supp} tracks whether later context needed to apply it survives; \textbf{Behav} tracks whether the final symbolic answer is consistent with both. Load, residue, and budget-hit metrics distinguish premise preservation from simply storing everything.

The diagnostics are organized by claim. Premise-realization and compression tests probe post-admission premise loss; heavy-pressure, threshold, and capacity sweeps test the omission--saturation tradeoff; and gate--strength, confidence$\times$strength, strength-source, and implicit-heuristic diagnostics test whether the proposed components remain distinguishable. The controlled regimes treat strength as an interface signal: they test how lifecycle control uses retention consequence once an approximate signal is available, while open-ended strength estimation is separated into coarse/noisy diagnostics and future work.

\section{Results}

\subsection{RQ1: Successful admission does not guarantee premise realization}

This diagnostic rules out a simple write-time explanation for premise loss. All controllers first admit the high-consequence premise; only later do lower-consequence items create compression pressure. The final query requires both the original premise and later support context, so success requires the premise to remain operative rather than merely having been written once.

\begin{table}[t]
\centering
\footnotesize
\setlength{\tabcolsep}{3pt}
\begin{adjustbox}{max width=\columnwidth}
\begin{tabular}{@{}lccccccc@{}}
\toprule
Method & Adm & Prem & Supp & Behav & CritLoss & FinalLoad & BudgetHit \\
\midrule
StageMem & 1.00 & 1.00 & 1.00 & 1.00 & 0.00 & 0.6016 & 0.00 \\
Confidence-only & 1.00 & 0.00 & 1.00 & 0.00 & 1.00 & 0.8238 & 0.0861 \\
Single-state & 1.00 & 0.00 & 0.4144 & 0.00 & 1.00 & 0.1762 & 0.00 \\
Single-layer & 1.00 & 1.00 & 1.00 & 1.00 & 0.00 & 1.0000 & 0.5267 \\
\bottomrule
\end{tabular}
\end{adjustbox}
\caption{Premise-realization compression test. \textbf{Adm}: premise admission rate. \textbf{Prem}: premise survival/realization at final query time. \textbf{Supp}: late support-context realization. \textbf{Behav}: final premise-conditioned symbolic behavior. \textbf{CritLoss}: critical compression loss. \textbf{FinalLoad}: final normalized memory load. \textbf{BudgetHit}: memory-budget-hit rate.}
\label{tab:premise-realization-rewrite}
\end{table}

Table~\ref{tab:premise-realization-rewrite} separates admission, support retention, and premise realization. All methods admit the premise. Confidence-only settlement keeps the later support but loses the governing premise, so carried validity/support evidence is not enough. Single-state memory also loses the premise, while single-layer memory preserves the premise--support chain only at saturated load and with frequent budget hits. StageMem preserves the premise, support, and final behavior without falling back to saturated storage.

A lossy summarization-cycle abstraction in Appendix~\ref{sec:compression-cycle} shows that the failure is not limited to one-item eviction: recency, frequency, and confidence-only summaries can drop the governing premise under repeated compression.

\subsection{RQ2: Front-door omission and flat saturation are opposite failures}

The heavy admitted-content regime targets two opposite alternatives: admit conservatively at the front door, or retain broadly in a flat store. Many items are plausible at write time, but only one becomes important later.

\begin{table*}[t]
\centering
\small
\begin{adjustbox}{max width=\textwidth}
\begin{tabular}{lccccc}
\toprule
Method & Recall & ImpRet & FinalLoad & NonCrit & BudgetHit \\
\midrule
StageMem & 1.00 & 1.00 & 0.3077 & 0.7500 & 0.00 \\
Single-layer & 1.00 & 1.00 & 1.0000 & 0.7692 & 0.40 \\
MemGPT-style & 1.00 & 1.00 & 0.6923 & 0.7778 & 0.04 \\
HiMem-style & 1.00 & 1.00 & 0.5385 & 0.7500 & 0.00 \\
Mem0-style & 0.00 & 0.00 & 0.0769 & 0.0000 & 0.00 \\
MemoryBank-style & 0.00 & 0.00 & 0.0769 & 0.0000 & 0.00 \\
\bottomrule
\end{tabular}
\end{adjustbox}
\caption{Heavy admitted-content comparison across representative memory-control families using metrics defined for all control styles. Named baselines are shared-harness analogues that preserve dominant control principles rather than product reproductions. \textbf{ImpRet}: important-item retention. \textbf{FinalLoad}: final normalized memory load. \textbf{NonCrit}: noncritical residue. Hierarchical analogues preserve the target at different load levels, while conservative front-door and reinforced-flat analogues miss late-important content under this operating point.}
\label{tab:family-heavy-rewrite}
\end{table*}

\begin{figure*}[t]
\centering
\begin{tikzpicture}
\begin{axis}[
    name=threshold,
    width=0.48\textwidth,
    height=0.37\textwidth,
    xmin=0.43, xmax=0.97,
    ymin=-0.05, ymax=1.08,
    xlabel={Create threshold $\tau$},
    ylabel={Score},
    grid=both,
    grid style={line width=.1pt, draw=gray!25},
    major grid style={line width=.2pt,draw=gray!35},
    tick align=outside,
    legend style={at={(0.5,1.03)},anchor=south,draw=none,fill=white,fill opacity=0.9,text opacity=1,font=\scriptsize},
    every axis plot/.append style={line width=0.9pt, mark size=1.9pt},
]
\addplot+[mark=o] coordinates {
    (0.45,0.00) (0.55,0.00) (0.65,0.00) (0.72,0.00) (0.75,0.00)
    (0.78,0.00) (0.80,0.00) (0.82,0.00) (0.84,1.00) (0.85,1.00)
    (0.88,1.00) (0.90,1.00) (0.95,1.00)
};
\addlegendentry{Important-item retention}
\addplot+[mark=square*, dashed] coordinates {
    (0.45,0.0000) (0.55,0.0769) (0.65,0.0769) (0.72,0.0769) (0.75,0.0769)
    (0.78,0.0769) (0.80,0.0769) (0.82,0.0769) (0.84,1.0000) (0.85,1.0000)
    (0.88,1.0000) (0.90,1.0000) (0.95,1.0000)
};
\addlegendentry{Final load}
\end{axis}

\begin{axis}[
    name=pareto,
    at={(threshold.east)},
    xshift=0.07\textwidth,
    anchor=west,
    width=0.48\textwidth,
    height=0.37\textwidth,
    xmin=-0.03, xmax=1.05,
    ymin=-0.05, ymax=1.08,
    xlabel={Final normalized memory load},
    ylabel={Important-item retention},
    grid=both,
    grid style={line width=.1pt, draw=gray!25},
    major grid style={line width=.2pt,draw=gray!35},
    tick align=outside,
    legend style={at={(0.5,1.03)},anchor=south,draw=none,fill=white,fill opacity=0.9,text opacity=1,font=\scriptsize},
    every axis plot/.append style={line width=0.9pt},
]
\addplot+[mark=o, mark size=2.2pt] coordinates {(0.0000,0.00) (0.0769,0.00) (1.0000,1.00)};
\addlegendentry{Unique front-door operating points}
\addplot+[only marks, mark=*, mark size=3.0pt] coordinates {(0.3077,1.00)};
\addlegendentry{StageMem}
\addplot+[only marks, mark=square*, mark size=3.0pt] coordinates {(1.0000,1.00)};
\addlegendentry{Single-layer}
\end{axis}
\end{tikzpicture}
\caption{Front-door threshold sweep and Pareto view for the heavy admitted-content regime. The left panel shows the threshold sweep directly; the right panel collapses repeated thresholds into unique load--retention operating points and overlays StageMem and the single-layer anchor. Threshold tuning moves the front-door-only controller between omission and saturated retention without reaching the moderate-load full-retention region occupied by StageMem.}
\label{fig:front-door-pareto-rewrite}
\end{figure*}

Table~\ref{tab:family-heavy-rewrite} shows the split. Conservative front-door-style analogues remain light but miss late-important content. Single-layer memory preserves the target by filling the store. Hierarchical and tiered analogues can preserve the target at different load levels. StageMem occupies the intended operating region: retaining the important item without moving to the saturated flat-memory point.

Figure~\ref{fig:front-door-pareto-rewrite} shows that this is not just a threshold choice: sweeping the front-door threshold moves between omission and saturated retention rather than reaching the moderate-load full-retention region. Capacity scaling in Figure~\ref{fig:pressure-scaling-rewrite} shows the same pattern across budgets.

\begin{figure}[t]
\centering
\begin{tikzpicture}
\begin{axis}[
    width=0.98\columnwidth,
    height=0.74\columnwidth,
    symbolic x coords={Very tight,Tight,Default,Roomy,Very roomy},
    xtick=data,
    ymin=0, ymax=1.08,
    ylabel={Final normalized memory load},
    xlabel={Capacity setting},
    grid=both,
    grid style={line width=.1pt, draw=gray!25},
    major grid style={line width=.2pt,draw=gray!35},
    tick align=outside,
    tick label style={font=\scriptsize},
    label style={font=\small},
    legend columns=2,
    legend style={at={(0.5,1.03)},anchor=south,draw=none,fill=white,fill opacity=0.9,text opacity=1,font=\scriptsize},
    every axis plot/.append style={line width=0.9pt, mark size=2.0pt},
]
\addplot+[mark=*] coordinates {(Very tight,0.2857) (Tight,0.3000) (Default,0.3077) (Roomy,0.3889) (Very roomy,0.4545)};
\addlegendentry{StageMem}
\addplot+[mark=square*] coordinates {(Very tight,1.0000) (Tight,1.0000) (Default,1.0000) (Roomy,1.0000) (Very roomy,0.8636)};
\addlegendentry{Single-layer}
\addplot+[mark=triangle*] coordinates {(Very tight,1.0000) (Tight,1.0000) (Default,0.6923) (Roomy,0.8333) (Very roomy,0.6818)};
\addlegendentry{MemGPT-style}
\addplot+[mark=diamond*] coordinates {(Very tight,0.4286) (Tight,0.5000) (Default,0.5385) (Roomy,0.5556) (Very roomy,0.3182)};
\addlegendentry{HiMem-style}
\addplot+[mark=o] coordinates {(Very tight,0.1429) (Tight,0.1000) (Default,0.0769) (Roomy,0.0556) (Very roomy,0.0455)};
\addlegendentry{Mem0-style}
\end{axis}
\end{tikzpicture}
\caption{Capacity-based pressure scaling on the heavy admitted-content regime. StageMem keeps important-item retention stable while avoiding the saturated flat-memory operating point.}
\label{fig:pressure-scaling-rewrite}
\end{figure}

\subsection{RQ3: Validity evidence and retention consequence fail differently}

We next test whether confidence, strength, and lifecycle staging can substitute for one another. The diagnostic includes a valid high-consequence premise, later support, high-confidence local distractors, and low-confidence high-cue false premises. The point is to place both error directions in the same stream: false high-cue content should be rejected, while valid costly premises should survive.

\begin{table}[t]
\centering
\footnotesize
\setlength{\tabcolsep}{3pt}
\begin{adjustbox}{max width=\columnwidth}
\begin{tabular}{@{}lcccc@{}}
\toprule
Method & Prem & Supp & Behav & FalsePrem \\
\midrule
Front-door gate only & 0.00 & 1.00 & 0.00 & 0.00 \\
Confidence-only lifecycle & 0.02 & 0.00 & 0.00 & 0.00 \\
Strength-only lifecycle & 0.00 & 1.00 & 0.00 & 2.00 \\
Confidence+strength lifecycle & 1.00 & 1.00 & 1.00 & 0.00 \\
\bottomrule
\end{tabular}
\end{adjustbox}
\caption{Gate--strength--lifecycle diagnostic. \textbf{Prem}: valid high-consequence premise retained. \textbf{Supp}: support context retained. \textbf{Behav}: final symbolic behavior is correct and not governed by false premises. \textbf{FalsePrem}: average protected false high-strength-looking premises.}
\label{tab:gate-strength-lifecycle}
\end{table}

Table~\ref{tab:gate-strength-lifecycle} shows complementary failures. A front-door gate keeps support but loses the old premise. Confidence-only lifecycle control lacks a forgetting-cost state. Strength-only lifecycle control protects false high-cue content without a validity gate. The combined lifecycle preserves the true premise and support while rejecting spurious high-strength cues.

\begin{table*}[t]
\centering
\scriptsize
\begin{adjustbox}{max width=\textwidth}
\begin{tabular}{lccccccccc}
\toprule
Method & HcHs & McHs & HcLs & LcLs & GoodProt & BadProt & LocalProt & ProtLoad & Load \\
\midrule
StageMem & 1.00 & 1.00 & 0.00 & 0.00 & 1.00 & 0.00 & 0.00 & 2.00 & 7.00 \\
Cue-aware flat score & 1.00 & 1.00 & 0.00 & 0.00 & 1.00 & 0.00 & 5.00 & 7.00 & 7.00 \\
Binary strength flag & 1.00 & 1.00 & 0.00 & 0.00 & 1.00 & 0.00 & 0.00 & 2.00 & 2.00 \\
Confidence-only staged & 0.00 & 0.00 & 0.03 & 0.00 & 0.00 & 0.02 & 3.00 & 3.00 & 5.00 \\
Confidence flat & 0.01 & 0.00 & 0.64 & 0.00 & 0.00 & 0.32 & 6.99 & 7.00 & 7.00 \\
\bottomrule
\end{tabular}
\end{adjustbox}
\caption{Confidence$\times$strength quadrant diagnostic. \textbf{HcHs}: high-confidence/high-strength item. \textbf{McHs}: medium-confidence/high-strength item. \textbf{HcLs}: high-confidence/low-strength item. \textbf{LcLs}: low-confidence/low-strength item. \textbf{GoodProt}: average protected retention of HcHs and McHs. \textbf{BadProt}: average protected retention of HcLs and LcLs. \textbf{LocalProt}: protected local distractors.}
\label{tab:quadrant-diagnostic-rewrite}
\end{table*}

The confidence$\times$strength diagnostic in Table~\ref{tab:quadrant-diagnostic-rewrite} shows the boundary more directly. Binary strength succeeds in this clean quadrant setting, but pressure and bridge diagnostics expose the harder case where many items can plausibly be flagged important. Confidence-only systems protect high-confidence local distractors while missing high-strength targets.

The strength-source sensitivity in Table~\ref{tab:appendix-strength-source} further varies the source of the strength signal. Noisy and coarse cue-derived strength preserve the premise--support chain, while a generic-importance proxy misses needed support. This supports the interface reading: the controller does not require exact task-side labels, but it does require a consequence signal that remains actionable under settlement. Together, these diagnostics show that the result is not explained by a single item score.

\subsection{RQ4: Implicit heuristics recover partial signal but not clean control}
\label{sec:implicit-heuristics}

The strongest heuristic objection is that recency, frequency, query relevance, or generic importance may already approximate retention consequence. We test this with a stream where each heuristic recovers one useful part of memory but conflicts with another.

\begin{table}[t]
\centering
\footnotesize
\setlength{\tabcolsep}{3pt}
\begin{adjustbox}{max width=\columnwidth}
\begin{tabular}{lccccc}
\toprule
Method & Prem & Supp & Behav & FalsePrem & Residue \\
\midrule
Recency priority & 0.00 & 0.00 & 0.00 & 1.00 & 6.00 \\
Frequency priority & 0.02 & 1.00 & 0.02 & 0.00 & 5.98 \\
Query relevance priority & 0.00 & 1.00 & 0.00 & 0.02 & 5.98 \\
Generic importance priority & 1.00 & 1.00 & 0.00 & 4.00 & 1.00 \\
Cue-aware flat score & 1.00 & 1.00 & 0.00 & 4.00 & 1.00 \\
Explicit strength lifecycle & 1.00 & 1.00 & 1.00 & 0.00 & 0.00 \\
\bottomrule
\end{tabular}
\end{adjustbox}
\caption{Implicit retention heuristic diagnostic. Recency, frequency, query relevance, and generic-importance proxies each recover part of the stream, but each confounds retention consequence with another signal. The lifecycle row uses explicit strength for retention consequence and confidence as a validity gate. \textbf{Prem}: valid premise retained. \textbf{Supp}: support retained. \textbf{Behav}: final behavior correct. \textbf{Residue}: protected local distractors.}
\label{tab:implicit-heuristic}
\end{table}

Table~\ref{tab:implicit-heuristic} shows that the heuristics fail in different ways. Recency follows recent residue. Frequency and query relevance preserve later support but lose the sparse governing premise. Generic importance and cue-aware flat scoring retain premise and support, but also protect false high-importance-looking premises. The explicit-strength lifecycle is the only setting in this diagnostic that retains premise and support while filtering false premises through confidence. The narrow takeaway is that retention consequence may correlate with familiar heuristics, but leaving it implicit entangles it with signals that do different control work.

\subsection{RQ5: External tasks check compatibility outside the diagnostic regimes}

External adaptations check compatibility with retrieval-heavy settings rather than isolate the target failure. Corrected HotpotQA is used as a graph-compatibility probe: both graph variants saturate relation accuracy, while StageMem-G slightly lowers load and write amplification under the shared harness. LoCoMo and QMSum adaptations are near-saturated and are summarized textually in the appendix as coverage checks. These results support compatibility with relation-heavy retrieval and long-document coverage; the controlled regimes remain the mechanism evidence.

\paragraph{Summary of failure axes.}
Across diagnostics, the failures align with distinct control axes. Front-door control fails by omission; confidence-only control can preserve credible residue while losing costly premises; strength-only control can over-protect false high-cue content; and flat or binary-priority controls tend toward saturation or brittle tie-breaking once capacity binds. Successful hierarchical or priority-aware variants show that the core variable can be instantiated across different shared-harness control styles; the controlled failures arise when the signal is collapsed into a flat, ungated, or non-actionable priority under maintenance pressure.

\section{Discussion}

The main contribution is to turn retention consequence into a maintained control state. The system needs a state variable that remains available when admitted content is later summarized, promoted, compressed, or evicted. The experiments instantiate this variable through staged memory, priority-aware summarization, and simpler flags in clean settings. StageMem provides one inspectable version in which the variable is graded, validity-gated, and connected to lifecycle settlement.

In a deployed memory system, this lifecycle-control layer would sit between extraction and generation. An extractor proposes candidate memory items; verification, contradiction detection, source reliability, or local support initialize and update confidence; cue rules, user pinning, learned salience, feedback, or dependency structure initialize and update strength; lifecycle settlement decides what remains transient, becomes working memory, or is made durable; retrieval then exposes selected memory to the generator. The exact substrate may be a summary, graph, vector store, database, or hybrid system. The abstraction here is not tied to one backend: it specifies which state should remain available when admitted content is later compressed, promoted, or forgotten.

We use controlled diagnostics as the primary evidence because the contribution is a control abstraction. Given the same admitted stream, budget pressure, and readout, they make visible what changes when retention consequence is carried forward as state. External adaptations check relation traversal, retrieval, and coverage compatibility, while the de-identified natural-language bridge asks whether the same premise-versus-local-residue tradeoff appears in coding-assistant histories. Product-scale benchmarks are a natural next layer once the control variable is defined.

\section{Conclusion}

We studied persistent memory as a post-admission lifecycle-control problem. The central failure is that an admitted premise can later lose operational force under maintenance pressure. This requires separating carried validity/support evidence from retention consequence: confidence tracks whether an item remains usable, while strength tracks how costly it would be to lose.

StageMem instantiates this separation in a transient/working/durable controller, where confidence and strength act during promotion, compression, and eviction. Controlled diagnostics show that explicit retention consequence changes the failure surface, preserving costly premises and needed support without relying on conservative omission, false high-cue protection, or saturated retention. Lifecycle-managed memory offers a reusable lens for systems where the key question is not only what to write, but how admitted content remains actionable under pressure.

\section*{Limitations}

This paper fixes several layers of the memory stack in order to study lifecycle control directly. Candidate proposal, symbolic readout, and the initial consequence cues are controlled, while the maintenance policy varies. This makes the results diagnostic rather than product-level performance estimates. Baseline implementations are control-style approximations within a shared harness rather than exact reproductions of full memory systems, and the external adaptations check compatibility along retrieval, relation, and coverage axes.

The current experiments use simple auditable strength initializers to focus on how retention consequence is used during maintenance. Strength estimation itself is an interface: richer estimators could be learned from interaction history, user feedback, task dependencies, or domain structure and plugged into the same lifecycle controller. The positive results with noisy and coarse cue-derived signals suggest that the mechanism does not require a highly optimized estimator to expose the targeted failure modes, but product-scale estimation remains an important next step. Better calibration of confidence under contradiction, source uncertainty, and stale evidence is also left open. Broader natural-language multi-cycle benchmarks would be valuable once they can measure lifecycle failure without collapsing it into extraction, compression, retrieval, generation, or judging errors.

\section*{Ethics Statement}

Persistent memory can store sensitive user information, shape long-term model behavior, and create trust expectations. Lifecycle-managed memory should therefore be paired with deletion controls, user-facing transparency, and careful handling of privacy-sensitive content. This paper studies controlled memory-management mechanisms and does not advocate unrestricted retention of user data.

\bibliography{custom}

\FloatBarrier
\appendix

\section{Shared Harness and Strength Sources}

The named baselines are shared-harness analogues rather than full reproductions of product systems. Mem0-style is implemented as a conservative front-door controller; MemGPT-style as aggressive hierarchical retention; HiMem-style as hybrid layered retention; and MemoryBank-style as reinforced flat storage. This keeps the comparison focused on post-admission control behavior under matched streams, budgets, retrieval interfaces, and metrics. Product-specific extraction stacks, storage substrates, graph-native execution, tool-use loops, and personalization pipelines are outside this shared comparison contract.

We use different strength sources for different diagnostic roles. The main premise-realization and pressure regimes use task-side strength as a mechanism probe: they test whether explicit retention consequence helps once the relevant consequence signal is available. Component ablations use fixed or removed strength to show what fails when retention consequence is collapsed or badly calibrated. Cue-aware flat, binary-flag, and front-door sweeps are deliberately allowed to use simple cue-derived priority signals, so that the comparison tests lifecycle use of the signal rather than access to the cue itself. The natural-language bridge uses a fixed cue-to-strength rule over explicit constraint language, not per-item oracle labels. The strength-source sensitivity diagnostic adds noisy, coarse cue-derived, and generic-importance proxy sources to distinguish lifecycle use of retention consequence from exact task-side labeling. Learned estimation of retention consequence from raw interaction is left to future work.

\section{Additional Controlled Results}

\paragraph{Premise realization.}
The main paper reports the compact premise-realization comparison. Table~\ref{tab:appendix-premise-realization} gives the fuller score matrix, and Table~\ref{tab:appendix-premise-realization-sensitivity} reports the main sensitivity checks. These checks preserve the qualitative reading: the behavior is not a monotone consequence of increasing all salience-like signals, and calibration of validity/support evidence remains important.

\begin{table*}[!t]
\centering
\scriptsize
\begin{adjustbox}{max width=\textwidth}
\begin{tabular}{lcccccccc}
\toprule
Method & Adm & Prem & Supp & Behav & CritLoss & FinalLoad & NonCrit & BudgetHit \\
\midrule
StageMem & 1.00 & 1.00 & 1.00 & 1.00 & 0.00 & 0.6016 & 0.3694 & 0.0000 \\
Confidence-only settlement & 1.00 & 0.00 & 1.00 & 0.00 & 1.00 & 0.8238 & 0.6744 & 0.0861 \\
Single-state memory & 1.00 & 0.00 & 0.4144 & 0.00 & 1.00 & 0.1762 & 0.6305 & 0.0000 \\
Single-layer & 1.00 & 1.00 & 1.00 & 1.00 & 0.00 & 1.0000 & 0.6207 & 0.5267 \\
\bottomrule
\end{tabular}
\end{adjustbox}
\caption{Premise-realization compression full matrix using cross-variant metrics. \textbf{Adm}: premise admission rate. \textbf{Prem}: premise survival/realization rate at final query time. \textbf{Supp}: late support-context realization rate. \textbf{Behav}: final premise-conditioned symbolic behavior realization rate. \textbf{CritLoss}: critical compression loss rate. \textbf{FinalLoad}: final normalized memory load. \textbf{NonCrit}: noncritical item residue. \textbf{BudgetHit}: memory-budget-hit rate.}
\label{tab:appendix-premise-realization}
\end{table*}

\begin{table*}[!t]
\centering
\scriptsize
\begin{adjustbox}{max width=\textwidth}
\begin{tabular}{llccccc}
\toprule
Analysis & Setting & Prem & Supp & Behav & CritLoss & FinalLoad \\
\midrule
\multirow{3}{*}{Shock scale}
& $0.50\times$ & 1.00 & 1.00 & 1.00 & 0.00 & 0.6120 \\
& $1.00\times$ & 1.00 & 1.00 & 1.00 & 0.00 & 0.6120 \\
& $1.50\times$ & 1.00 & 1.00 & 1.00 & 0.00 & 0.6120 \\
\midrule
\multirow{3}{*}{Confidence scale}
& $0.70\times$ & 0.00 & 0.3264 & 0.00 & 1.00 & 0.1658 \\
& $0.85\times$ & 0.00 & 1.00 & 0.00 & 1.00 & 0.3898 \\
& $1.00\times$ & 1.00 & 1.00 & 1.00 & 0.00 & 0.6120 \\
\midrule
\multirow{4}{*}{Fixed init strength}
& $0.02$ & 1.00 & 1.00 & 1.00 & 0.00 & 0.3898 \\
& $0.10$ & 1.00 & 1.00 & 1.00 & 0.00 & 0.6120 \\
& $0.20$ & 0.00 & 1.00 & 0.00 & 1.00 & 0.8342 \\
& $0.30$ & 0.00 & 1.00 & 0.00 & 1.00 & 0.8342 \\
\bottomrule
\end{tabular}
\end{adjustbox}
\caption{StageMem sensitivity analysis on the premise-realization regime. Shock rescaling preserves the main result, while lowered confidence and overly high fixed initial strength expose calibration sensitivity. \textbf{Prem}: premise survival. \textbf{Supp}: support realization. \textbf{Behav}: final behavior realization.}
\label{tab:appendix-premise-realization-sensitivity}
\end{table*}

\paragraph{Strength-source sensitivity.}
Table~\ref{tab:appendix-strength-source} tests the strength interface under increasingly approximate sources. The goal is to separate two questions: whether retention consequence is useful as a lifecycle state, and whether this paper solves open-ended strength estimation. Noisy and coarse cue-derived strength preserve the premise--support chain when paired with confidence gating, while a generic-importance proxy misses needed support and flat generic importance protects false high-cue premises. The result supports the interface reading used in the paper: retention consequence can be supplied by different estimators, but it must remain explicit, validity-gated, and actionable during settlement.

\begin{table*}[!t]
\centering
\scriptsize
\begin{adjustbox}{max width=\textwidth}
\begin{tabular}{lccccccc}
\toprule
Method & Prem & Supp & Behav & FalsePrem & LocalResidue & ProtLoad & Load \\
\midrule
Flat generic-importance control & 1.00 & 1.00 & 0.00 & 4.00 & 1.00 & 7.00 & 7.00 \\
Lifecycle oracle strength & 1.00 & 1.00 & 1.00 & 0.00 & 0.00 & 2.00 & 7.00 \\
Lifecycle noisy strength & 1.00 & 1.00 & 1.00 & 0.00 & 0.00 & 2.00 & 7.00 \\
Lifecycle coarse cue strength & 1.00 & 1.00 & 1.00 & 0.00 & 0.00 & 2.00 & 7.00 \\
Lifecycle generic-importance proxy & 1.00 & 0.01 & 0.01 & 0.00 & 0.00 & 1.01 & 6.01 \\
\bottomrule
\end{tabular}
\end{adjustbox}
\caption{Strength-source sensitivity diagnostic. All lifecycle rows use the same confidence gate and settlement rules but vary the source of the strength signal. \textbf{Noisy} adds bounded random perturbation to the task-side signal; \textbf{coarse cue} uses simple role-level cue buckets; \textbf{generic-importance proxy} uses generic importance and confidence without access to task consequence. \textbf{FalsePrem}: protected false premises. \textbf{ProtLoad}: protected working/durable load.}
\label{tab:appendix-strength-source}
\end{table*}

\paragraph{Heavy pressure.}
Table~\ref{tab:appendix-heavy-family} gives the full metric matrix for the heavy admitted-content regime. Table~\ref{tab:appendix-pressure-scaling} reports the capacity sweep underlying the operating-point figure in the main text. Additional sample-count and episode-length sweeps are included in the supplementary files and show the same qualitative pattern: conservative front-door analogues remain light but miss late-important content, flat retention saturates, and staged or hierarchical controls occupy different load points.

\begin{table*}[!t]
\centering
\scriptsize
\resizebox{\textwidth}{!}{%
\begin{tabular}{lcccccccccc}
\toprule
Method & Recall & ImpRet & FinalLoad & NonCrit & BudgetHit & Regret & Useful & WriteAmp & Hit@1 & MRR \\
\midrule
StageMem & 1.00 & 1.00 & 0.3077 & 0.7500 & 0.00 & 0.2500 & 0.2500 & 15.00 & 1.00 & 1.00 \\
Single-layer & 1.00 & 1.00 & 1.0000 & 0.7692 & 0.40 & 0.0000 & 0.2308 & 8.00 & 1.00 & 1.00 \\
Mem0-style & 0.00 & 0.00 & 0.0769 & 0.0000 & 0.00 & 0.0000 & 1.0000 & 0.3333 & 0.00 & 0.00 \\
MemGPT-style & 1.00 & 1.00 & 0.6923 & 0.7778 & 0.04 & 0.0000 & 0.2222 & 19.3333 & 1.00 & 1.00 \\
HiMem-style & 1.00 & 1.00 & 0.5385 & 0.7500 & 0.00 & 0.2195 & 0.2500 & 8.00 & 1.00 & 1.00 \\
MemoryBank-style & 0.00 & 0.00 & 0.0769 & 0.0000 & 0.00 & 0.0000 & 1.0000 & 0.6667 & 0.00 & 0.00 \\
\bottomrule
\end{tabular}
}
\caption{Heavy admitted-content family full matrix using cross-family metrics. \textbf{Recall}: heavy-regime recall accuracy. \textbf{ImpRet}: important-item retention. \textbf{FinalLoad}: final normalized memory load. \textbf{NonCrit}: noncritical item residue. \textbf{BudgetHit}: memory-budget-hit rate. \textbf{Regret}: eviction regret. \textbf{Useful}: retained useful ratio. \textbf{WriteAmp}: write amplification. \textbf{Hit@1}: retrieval hit@1.}
\label{tab:appendix-heavy-family}
\end{table*}

\begin{table}[t]
\centering
\scriptsize
\setlength{\tabcolsep}{3pt}
\begin{adjustbox}{max width=\columnwidth}
\begin{tabular}{llccc}
\toprule
Capacity & Method & ImpRet & FinalLoad & NonCrit \\
\midrule
\multirow{5}{*}{Very tight}
& StageMem & 1.00 & 0.2857 & 0.5000 \\
& Single-layer & 1.00 & 1.0000 & 0.5714 \\
& Mem0-style & 0.00 & 0.1429 & 0.0000 \\
& MemGPT-style & 1.00 & 1.0000 & 0.7143 \\
& HiMem-style & 1.00 & 0.4286 & 0.5000 \\
\midrule
\multirow{5}{*}{Tight}
& StageMem & 1.00 & 0.3000 & 0.6667 \\
& Single-layer & 1.00 & 1.0000 & 0.7000 \\
& Mem0-style & 0.00 & 0.1000 & 0.0000 \\
& MemGPT-style & 1.00 & 1.0000 & 0.8000 \\
& HiMem-style & 1.00 & 0.5000 & 0.6667 \\
\midrule
\multirow{5}{*}{Default}
& StageMem & 1.00 & 0.3077 & 0.7500 \\
& Single-layer & 1.00 & 1.0000 & 0.7692 \\
& Mem0-style & 0.00 & 0.0769 & 0.0000 \\
& MemGPT-style & 1.00 & 0.6923 & 0.7778 \\
& HiMem-style & 1.00 & 0.5385 & 0.7500 \\
\midrule
\multirow{5}{*}{Roomy}
& StageMem & 1.00 & 0.3889 & 0.7143 \\
& Single-layer & 1.00 & 1.0000 & 0.8333 \\
& Mem0-style & 0.00 & 0.0556 & 0.0000 \\
& MemGPT-style & 1.00 & 0.8333 & 0.8667 \\
& HiMem-style & 1.00 & 0.5556 & 0.8333 \\
\midrule
\multirow{5}{*}{Very roomy}
& StageMem & 1.00 & 0.4545 & 0.7000 \\
& Single-layer & 1.00 & 0.8636 & 0.7895 \\
& Mem0-style & 0.00 & 0.0455 & 0.0000 \\
& MemGPT-style & 1.00 & 0.6818 & 0.8000 \\
& HiMem-style & 1.00 & 0.3182 & 0.7500 \\
\bottomrule
\end{tabular}
\end{adjustbox}
\caption{Capacity-based pressure scaling for the heavy admitted-content regime. \textbf{ImpRet}: important-item retention. \textbf{FinalLoad}: final normalized memory load. \textbf{NonCrit}: noncritical item residue.}
\label{tab:appendix-pressure-scaling}
\end{table}

\section{Compression-Cycle and Salience Details}
\label{sec:compression-cycle}

The lossy summarization-cycle abstraction checks that premise loss is not limited to one-item eviction: repeated context overflow can also remove a governing premise during summary maintenance. Flat salience succeeds in this clean abstraction, so the table is evidence for explicit priority under lossy compression rather than for the uniqueness of StageMem. The learned-salience table records the current salience initializer behavior.

\begin{table}[t]
\centering
\scriptsize
\setlength{\tabcolsep}{2.5pt}
\begin{adjustbox}{max width=\columnwidth}
\begin{tabular}{@{}lcccccc@{}}
\toprule
Summary policy & Prem & Supp & Behav & Residue & Details & LossCycle \\
\midrule
Strength-aware & 1.00 & 1.00 & 1.00 & 0.91 & 12.53 & 0.00 \\
Confidence-only & 0.00 & 0.00 & 0.00 & 1.00 & 4.00 & 1.00 \\
Flat salience & 1.00 & 1.00 & 1.00 & 0.90 & 11.51 & 0.00 \\
Frequency & 0.00 & 0.09 & 0.00 & 0.98 & 15.57 & 2.81 \\
Recency & 0.00 & 0.00 & 0.00 & 0.18 & 5.65 & 1.00 \\
\bottomrule
\end{tabular}
\end{adjustbox}
\caption{Lossy summarization-cycle abstraction. \textbf{Prem}: critical premise survives. \textbf{Supp}: enough late support survives. \textbf{Behav}: final premise-conditioned behavior can be realized. \textbf{Residue}: at least one local-scope instruction remains in the final summary. \textbf{Details}: retained detail count. \textbf{LossCycle}: average cycle at which the premise is first lost.}
\label{tab:appendix-summarization-cycle}
\end{table}

\begin{table}[t]
\centering
\small
\setlength{\tabcolsep}{3pt}
\begin{adjustbox}{max width=\columnwidth}
\begin{tabular}{@{}lcccccc@{}}
\toprule
Objective & PPrec & PRec & C$\to$R & R$\to$D & SecDis & FillFP \\
\midrule
Point target v2 & 0.3783 & 1.00 & 4.1445 & 1.1426 & 58 & 145 \\
Ranking-margin v2 & 0.4000 & 1.00 & 4.0000 & 1.0000 & 64 & 128 \\
\bottomrule
\end{tabular}
\end{adjustbox}
\caption{Learned salience full matrix. \textbf{PPrec}/\textbf{PRec}: promotion precision/recall. \textbf{C$\to$R}: cache-to-RAM promotions. \textbf{R$\to$D}: RAM-to-disk promotions. \textbf{SecDis}: supported-secondary early discards. \textbf{FillFP}: filler false promotions.}
\label{tab:appendix-salience}
\end{table}

\section{External Compatibility and Natural-Language Bridge}

External adaptations are compatibility checks. Corrected HotpotQA tests relation-heavy graph compatibility; LoCoMo and QMSum were near-saturated under the current symbolic harness and are therefore summarized textually. The natural-language bridge is a small realism sanity check over de-identified coding-assistant histories, not a public benchmark or end-to-end agent evaluation. It asks whether a premise-versus-local-residue tradeoff appears when all policies receive the same extracted candidates and strength is initialized by a fixed cue-to-strength rule rather than by per-item oracle labels.

\begin{table}[t]
\centering
\small
\setlength{\tabcolsep}{4pt}
\begin{adjustbox}{max width=\columnwidth}
\begin{tabular}{lcccccc}
\toprule
Method & RelAcc & AvgLoad & Hit@1 & MRR & Useful & WriteAmp \\
\midrule
Flat graph & 1.00 & 0.0496 & 1.00 & 1.00 & 0.1503 & 6.9565 \\
StageMem-G & 1.00 & 0.0410 & 1.00 & 1.00 & 0.1901 & 5.5652 \\
\bottomrule
\end{tabular}
\end{adjustbox}
\caption{Graph compatibility full matrix on corrected external HotpotQA. \textbf{RelAcc}: relation accuracy. \textbf{AvgLoad}: average normalized memory load. \textbf{WriteAmp}: write amplification.}
\label{tab:comparison-matrix-graph}
\end{table}

\begin{figure*}[!t]
\centering
\begin{tikzpicture}
\begin{axis}[
    name=retention,
    ybar,
    bar width=5.5pt,
    width=0.47\textwidth,
    height=0.35\textwidth,
    ymin=0.60, ymax=1.05,
    ylabel={Long-memory premise retention},
    symbolic x coords={9,13,25},
    xtick=data,
    xlabel={Capacity},
    enlarge x limits=0.22,
    grid=both,
    grid style={line width=.1pt, draw=gray!25},
    major grid style={line width=.2pt,draw=gray!35},
    tick align=outside,
    legend style={at={(0.5,1.03)},anchor=south,draw=none,fill=white,fill opacity=0.9,text opacity=1,font=\scriptsize},
    every axis plot/.append style={line width=0.4pt},
]
\addplot+[fill=gray!20, draw=black] coordinates {(9,0.8462) (13,0.8462) (25,1.0000)};
\addlegendentry{StageMem proxy}
\addplot+[fill=gray!50, draw=black] coordinates {(9,0.6923) (13,0.7692) (25,0.9231)};
\addlegendentry{Best clean baseline}
\addplot+[fill=gray!80, draw=black] coordinates {(9,1.0000) (13,1.0000) (25,1.0000)};
\addlegendentry{Unbounded binary flag}
\end{axis}

\begin{axis}[
    name=load,
    at={(retention.east)},
    xshift=0.08\textwidth,
    anchor=west,
    ybar,
    bar width=5.5pt,
    width=0.47\textwidth,
    height=0.35\textwidth,
    ymin=0, ymax=55,
    ylabel={Average retained items},
    symbolic x coords={9,13,25},
    xtick=data,
    xlabel={Capacity},
    enlarge x limits=0.22,
    grid=both,
    grid style={line width=.1pt, draw=gray!25},
    major grid style={line width=.2pt,draw=gray!35},
    tick align=outside,
    legend style={at={(0.5,1.03)},anchor=south,draw=none,fill=white,fill opacity=0.9,text opacity=1,font=\scriptsize},
    every axis plot/.append style={line width=0.4pt},
]
\addplot+[fill=gray!20, draw=black] coordinates {(9,8.08) (13,10.15) (25,22.00)};
\addlegendentry{StageMem proxy}
\addplot+[fill=gray!50, draw=black] coordinates {(9,8.54) (13,11.92) (25,21.15)};
\addlegendentry{Best clean baseline}
\addplot+[fill=gray!80, draw=black] coordinates {(9,50.62) (13,50.62) (25,50.62)};
\addlegendentry{Unbounded binary flag}
\end{axis}
\end{tikzpicture}
\caption{Natural-language history bridge sanity check. The left panel reports retention of curated long-memory premise candidates; the right panel reports average retained items. ``Best clean baseline'' is the best swept non-lifecycle baseline under matched capacity, zero over-budget rate, and zero local-control retention. The unbounded binary flag illustrates the hoarding limit: it can preserve marked premises, but exceeds realistic budgets in most sessions.}
\label{fig:appendix-history-bridge}
\end{figure*}

The bridge supports a narrower point than the controlled regimes. It does not evaluate a live generator. Instead, it checks the memory state that a final natural-language query would receive: whether long-memory premise candidates remain available, local-control negative items are dropped, and retained content stays within capacity. The supplementary file \texttt{natural\_language\_case\_studies.md} contains qualitative final-query cases illustrating the same mechanism in repository constraints, report formatting, dependency licensing, and itinerary compression.

\section{Operational Details}

This section records the implementation-level update and salience details used by the current experiments. These equations define one stable instantiation of the framework rather than a claim that the exact parameterization is unique.

\paragraph{Continuous update dynamics.}
Let $c$ and $m$ denote the stored confidence and strength, and let $\hat{c}$ denote incoming evidence confidence. The current mainline first computes
\begin{align}
 e_{\mathrm{in}} &= \frac{\exp(\alpha \hat{c})-1}{\exp(\alpha)-1}, \\
 r_{\mathrm{conf}} &= \mathrm{normexp}(\sigma \cdot \mathrm{sign}), \\
 r_{\mathrm{str}} &= \exp(-\beta m),
\end{align}
where $\mathrm{sign}\in\{+1,-1\}$ indicates whether new evidence supports or conflicts with the stored value. The plasticity is
\begin{equation}
 p = \mathrm{clip}\!\left(\lambda e_{\mathrm{in}} r_{\mathrm{conf}} r_{\mathrm{str}}, p_{\min}, 1\right),
\end{equation}
with smaller $\lambda$ in deeper stages. Content is interpolated in the default continuous case, while accepted same-anchor conflict revisions replace the stored value directly. Confidence updates use the same input factors with supportive evidence moving toward $1$ and conflicting evidence moving toward $0$.

Strength grows more conservatively:
\begin{align}
 b &= \mathrm{clip}\!\left(0.60\hat{c}+0.20\,\mathrm{shock},0,1\right), \\
 m' &= m + p\cdot 0.35\cdot b\cdot (1-m).
\end{align}
Here \texttt{shock} is a task-side signal for contextual importance, and $(1-m)$ makes strength growth saturating.

\paragraph{Learned salience initialization.}
For admitted item embedding $x$, contextual strength signal $u$, incoming confidence $\hat{c}$, and shock signal $z$, the salience scorer predicts
\begin{align}
 h &= \mathrm{MLP}([x;u;\hat{c};z]), \\
 s_{\mathrm{sal}} &= \sigma(h), \\
 m_0 &= m_{\min}+(m_{\max}-m_{\min})s_{\mathrm{sal}}.
\end{align}
Thus learned salience initializes retention depth within a bounded range rather than replacing admission or lifecycle settlement.

\end{document}